\documentclass{article}

\usepackage[preprint]{neurips_2023}

\usepackage[utf8]{inputenc} 
\usepackage[T1]{fontenc}    
\usepackage[hidelinks]{hyperref}       
\usepackage{url}            
\usepackage{booktabs}       
\usepackage{amsfonts}       
\usepackage{nicefrac}       
\usepackage{microtype}      
\usepackage{xcolor}         

\usepackage{microtype}
\usepackage{graphicx}
\usepackage{wrapfig}

\usepackage{booktabs}
\usepackage{textcomp}

\usepackage{amsmath,amssymb} 
\usepackage{color, colortbl}

\usepackage{multirow}
\usepackage{bbm}

\usepackage{tikz}
\usepackage{comment}

\usepackage{algorithm}
\usepackage{algorithmic}

\usepackage{caption}
\usepackage{subcaption}

\usepackage{pifont}%

\title{Exploring Data Augmentations on \\ Self-/Semi-/Fully- Supervised Pre-trained Models}

\author{%
  Shentong Mo$^1$~Zhun Sun$^{2}$\thanks{Corresponding author.}~~Chao Li$^3$\\
  $^1$Carnegie Mellon University, $^2$Tohoku University \\ $^3$Center for Advanced Intelligence Project (AIP), RIKEN 
}

\begin{document}

\maketitle

\begin{abstract}

Data augmentation has become a standard component of vision pre-trained models to capture the invariance between augmented views. 
In practice, augmentation techniques that mask regions of a sample with zero/mean values or patches from other samples are commonly employed in pre-trained models with self-/semi-/fully-supervised contrastive losses.
However, the underlying mechanism behind the effectiveness of these augmentation techniques remains poorly explored.
To investigate the problems, we conduct an empirical study to quantify how data augmentation affects performance. Concretely, we apply 4 types of data augmentations termed with \texttt{Random Erasing}, \texttt{CutOut}, \texttt{CutMix} and \texttt{MixUp} to a series of self-/semi-/fully- supervised pre-trained models. We report their performance on vision tasks such as image classification, object detection, instance segmentation, and semantic segmentation. We then explicitly evaluate the invariance and diversity of the feature embedding.
We observe that:
1) Masking regions of the images decreases the invariance of the learned feature embedding while providing a more considerable diversity.
2) Manual annotations do not change the invariance or diversity of the learned feature embedding.
3) The \texttt{MixUp} approach improves the diversity significantly, with only a marginal decrease in terms of the invariance.

\end{abstract}
\vspace{-0.5em}

\section{Introduction}
\vspace{-0.5em}

Recently, self-/semi-/fully- supervised contrastive learning has achieved promising performance in learning meaningful representations during pre-training.
Besides, the pre-trained models are successfully transferred to many downstream tasks, such as image classification, object detection, and instance segmentation. Terminologically, self-supervised contrastive learning refers to the pre-training without any labels introduced. While we term it as the semi-/fully- supervised contrastive learning when providing partial/all ground truths labels.

In the pure self-supervised configurations, data augmentations act as an essential component of self-supervised contrastive learning~\cite{chen2020simple,he2019moco,chen2020mocov2}. The algorithms are optimized to minimize the distance between different augmented views from the same sample (\textit{a.k.a.} the anchor), while pushing views from different samples (the contrastive ones) away from the anchor.
On the other hand, previous studies~\cite{chen2020simple} show that with a limited amount of labels introduced, semi-supervised contrastive learning achieves better performance in related downstream tasks. Furthermore, fully-supervised contrastive learning with all ground truths further boosts the performance~\cite{khosla2020sup}.

In practice, augmentation techniques that mask regions of a sample with zero/mean values or patches from other samples are commonly employed in semi-/fully- supervised (non-contrastive) learning. However, this family of augmentation techniques is not often applied in contrastive configurations, and the underlying mechanism behind the effectiveness of these augmentation techniques remains poorly explored. 
In this study, we implement 4 types of data augmentations termed with \texttt{Random Erasing}, \texttt{CutOut}, \texttt{CutMix} and \texttt{MixUp} to a series of self-/semi-/fully- supervised pre-trained models. We then conduct a numerical study to quantify how data augmentation affects performance.

\begin{wrapfigure}{r}{0.4\textwidth}
\vspace{-1.0em}
  \begin{center}
\includegraphics[width=0.4\textwidth]{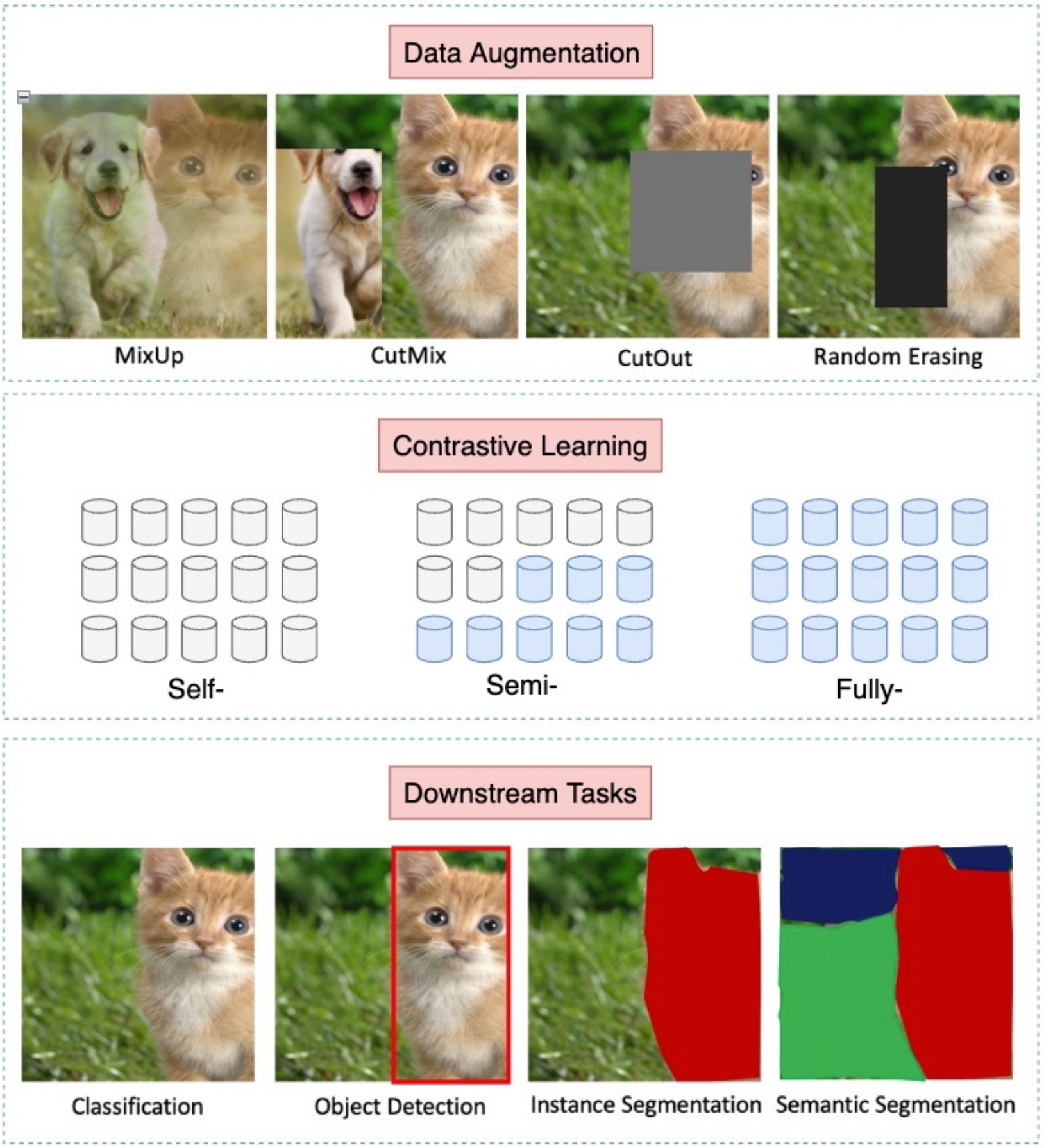}
\end{center}
  \caption{Illustration of our empirical study on four data augmentations (MixUp, CutMix, CutOut, Random Erasing), three pre-training types(self-, semi-, fully-supervised), and four downstream tasks (classification, object detection, instance segmentation, semantic segmentation).}
\label{fig: title_img}
\vspace{-1.0em}
\end{wrapfigure}

To this end, we clarify the terms \emph{invariance} and \emph{diversity} and provide the methods to calculate them explicitly. We then evaluate the invariance and diversity of the feature embedding of numerous pre-trained models. We demonstrate that \emph{invariance} and \emph{diversity} are closely related to the downstream tasks. Besides, we observe that:
1) Masking regions of the images decreases the invariance of the learned feature embedding while providing a more considerable diversity.
2) Manual annotations do not change the invariance or diversity of the learned feature embeddings.
3) The \texttt{MixUp} approach improves the diversity significantly, with only a marginal decrease in terms of the invariance.

Overall, the main contributions of this work can be summarized as follows:
\begin{itemize}
    \item We conduct a comprehensive empirical study by quantifying how data augmentation affects the self-/semi-/fully- supervised contrastive learning frameworks.
    \item We provide an approach to measure the quality of the augmented view by explicitly examining the invariance and diversity metrics for self-/semi-/fully- supervised pre-trained models. 
    \item Extensive experiments on various downstream benchmarks demonstrate that invariance and diversity are important metrics for the contrastive learning frameworks. Data augmentations that provide better invariance and diversity result in better performance in downstream tasks.
\end{itemize}

\vspace{-0.5em}

\section{Methodology}
\vspace{-0.5em}

In this work, we conduct an empirical study to quantify the effect of data augmentation techniques on the self-/semi-/fully- supervised contrastive learning frameworks.
First, we begin with the formal problem setup for this empirical study.
Then, we introduce self-/semi-/fully- supervised InfoNCE loss for comparisons.
Finally, we propose two metrics, invariance, and diversity, to measure the  quality of the augmented views between the anchor.

\noindent\textbf{Notations.}
Given a pre-training set of $N$ sample/label pairs, $\mathcal{N}=\{\boldsymbol{x}_i, \boldsymbol{y}_i\}_{k=1, \cdots, N}$.
Under the commonly-used contrastive learning setting~\cite{chen2020mocov2,he2019moco}, we generate two views $\boldsymbol{q}_i, \boldsymbol{k}_i$ for each sample $\boldsymbol{x}_i$. 
A set of negative samples for each sample $\boldsymbol{x}_i$ is $\mathcal{M}(i) = \{\boldsymbol{k}_m\}_{m =1, 2, \cdots, M}$ and $M$ is the number of negative samples.

\subsection{Preliminaries: Self- \& Fully-Supervised Contrastive Loss}

Under the self-supervised contrastive learning framework, the main objective for each sample $\boldsymbol{x}_i$ is to maximize the similarity between the query $\boldsymbol{q}_i$. The corresponding augmented view $\boldsymbol{k}_i$, while minimizing the similarity between the query $\boldsymbol{q}_i$ and the negative sample $\boldsymbol{k}_m$.
Thus, the overall objective $\mathcal{L}^{\mathrm{self}}$ is formulated as:
\begin{equation}\label{eq:self}
\begin{aligned}
    \mathcal{L}^{\mathrm{self}}  & = \sum_{i\in \mathcal{I}} \mathcal{L}_i^{\mathrm{self}} =  - \sum_{i\in \mathcal{I}} \log \frac{\kappa}{\kappa + \sum_{m\in \mathcal{M}(i)}\exp(\boldsymbol{q}_i\cdot\boldsymbol{k}_m /\tau)} 
\end{aligned}
\end{equation}
where $\kappa$ is the positive similarity term, $\exp(\mathbf{q}_i\cdot \mathbf{k}_{i} /\tau)$, and $\mathcal{M}(i)$ denote the set of negative samples.
$\tau$ is a temperature parameter.

By introducing all ground truths in the pre-training stage, we generate a new set $\mathcal{M}^\prime(i)$ of negative samples, where the labels of negative samples are different from that of the anchor.  
Then, we define the fully-supervised objective $\mathcal{L}^{\mathrm{full}}$ with the new negative set $\mathcal{M}^\prime(i)$ as:
\begin{equation}\label{eq:full}
\begin{aligned}
    \mathcal{L}^{\mathrm{full}} & = \sum_{i\in \mathcal{I}} \mathcal{L}_i^{\mathrm{full}} = \sum_{i\in \mathcal{I}} -\log \frac{\kappa}{\kappa + \sum_{m\in \mathcal{M}^\prime(i)}\exp(\boldsymbol{q}_i\cdot\boldsymbol{k}_m /\tau)}
\end{aligned}
\end{equation}
where $\mathcal{M}^\prime(i) = \{\boldsymbol{k}_m\vert\boldsymbol{y}_m \neq \boldsymbol{y}_i\}$, and other settings are the same as in Eq.~\ref{eq:self}.

\subsection{Semi-Supervised Contrastive Loss}

In practice, it is unrealistic to acquire all labels from a large-scale pre-training set. 
Instead, obtaining partial annotations is operable.
In this way, we split the original set $\mathcal{N}$ into two subsets, labelled set $\mathcal{D}$ and unlabelled set $\mathcal{U}$.
Given the sample $\boldsymbol{x}_i$ in the labelled set $\mathcal{D}$, we maintain a negative samples queue $\mathcal{M}_d(i)$ and a label queue $\mathcal{Y}_d(i)$.
In the meanwhile, we keep a negative samples queue $\mathcal{M}_u(i)$ for each sample in the unlabelled set $\mathcal{U}$.
Then, we apply the fully-supervised contrastive loss $\mathcal{L}_i^{\mathrm{full}}$ to the labelled set $\mathcal{D}$ and the self-supervised contrastive loss $\mathcal{L}_i^{\mathrm{self}}$ to the unlabelled set $\mathcal{U}$.
Therefore, the overall objective of semi-supervised contrastive loss is defined as: 

\begin{equation}
\begin{aligned}
    \mathcal{L}^{\mathrm{semi}} & = \sum_{i\in \mathcal{D}} \mathcal{L}_i^{\mathrm{full}} + \sum_{i\in \mathcal{U}} \mathcal{L}_i^{\mathrm{self}} = \sum_{i\in \mathcal{D}} -\log \frac{\kappa}{\kappa + \sum_{m\in \mathcal{M}_d(i)}\exp(\boldsymbol{q}_i\cdot\boldsymbol{k}_m /\tau)} \\
    & - \sum_{i\in \mathcal{U}} \log \frac{\kappa}{\kappa + \sum_{m\in \mathcal{M}_u(i)}\exp(\boldsymbol{q}_i\cdot\boldsymbol{k}_m /\tau)} 
\end{aligned}
\end{equation}

where $\mathcal{M}_d(i), \mathcal{M}_u(i)$ denotes the negative samples queue for the labelled set $\mathcal{D}$ and the unlabelled set $\mathcal{U}$. 
$\mathcal{M}_d(i) = \{\boldsymbol{k}_m\vert\boldsymbol{y}_d(i) \neq \boldsymbol{y}_i\}$.
Other terms are the same as in Eq.~\ref{eq:self} and ~\ref{eq:full}.

\subsection{Invariance}

In order to measure the invariance between the augmented views $\boldsymbol{q}_i$ and the anchor $\boldsymbol{x}_i$, we propose a metric to calculate the normalized similarity invariance of the views in terms of the embedding space.
Specifically, we take a set $\mathcal{V}_i$ of views, $\mathcal{V}_i = \{\boldsymbol{q}^v_i, v=1,\cdots, V\}$, by applying data augmentations to the original sample $\boldsymbol{x}_i$.
Then we calculate the normalized embeddings similarity between the augmented views $\boldsymbol{q}^v_i$ and the raw sample $\boldsymbol{x}_i$.
Thus, we formulate the invariance metric of augmented views as:

\begin{equation}
    \mathcal{L}_{inv} = \dfrac{1}{NV}\sum_{i=1}^N\sum_{v=1}^V \dfrac{\mathcal{S}(\boldsymbol{q}_i^v, \boldsymbol{x}_i)}{\mathcal{S}(\boldsymbol{x}_i,\boldsymbol{x}_i)}
\end{equation}
where $\mathcal{S}(\boldsymbol{x}_i, \boldsymbol{q}^v_i)$ denotes the dot product metric for calculating the distance between $\boldsymbol{q}^v_i$ and $\boldsymbol{x}_i$.
Note that $\mathcal{L}_{inv}$ achieves the maximum value $1$ when $\boldsymbol{q}_i^v = \boldsymbol{x}_i$.
This means that the augmented views have the maximum invariance from the anchor.

\subsection{Diversity}

In order to measure the quality of the augmented view in a comprehensive manner, we also propose to qualify the diversity of the augmented views.
Specifically, we introduce a metric named \textit{diversity} to measure how different the augmented views in the set $\mathcal{V}_i$ are.
Based on the dot product distance metric $\mathcal{S}$, we define the diversity between two augmented views $\boldsymbol{q}_i^v$ and $\boldsymbol{q}_i^w$ as: 

\begin{equation}
    \mathcal{L}_{div} = \dfrac{1}{NV(V-1)}\sum_{i=1}^N\sum_{v=1}^V\sum_{w\neq v}^V \exp \Big(\dfrac{\mathcal{S}(\boldsymbol{q}_i^v,\boldsymbol{q}^{w}_i)}{\sigma}\Big)
\end{equation}
where  $\mathcal{S}(\boldsymbol{q}_i^v,\boldsymbol{q}^{w}_i)$ denotes the dot product distance metric between $\boldsymbol{q}_i^v$ and $\boldsymbol{q}_i^w$.
$\sigma$ is a scale parameter.
In this way, we simultaneously maximize the diversity and invariance of the augmented views together to acquire views with best quality for self-/semi-/fully- supervised contrastive learning.

\begin{table*}[t]
	\renewcommand\tabcolsep{3.0pt}
	\centering
			\caption{Comparisons of linear classification evaluation on ImageNet-100 via applying four data augmentations to MoCo v2, where models are trained on frozen features from pre-trained encoders. Bold and underlined numbers denote the first and second place.} 
   \vspace{0.5em}
	\scalebox{0.85}{
		\begin{tabular}{lccccccccc}
			\toprule
			Method & Arch. & Param.(M) & Batch & Epochs & Top-1(\%) &  Top-5(\%) & $\mathcal{L}_{inv}$ & $\mathcal{L}_{div}$ \\
			\midrule
			MoCo v2~\cite{chen2020mocov2} & ResNet-50 & 24 & 256 & 200 & 81.65 & 95.77 & \textbf{0.72} & 0.23 \\  \hline
			MoCo v2 + Random Erasing & ResNet-50 & 24 & 256 & 200 & 81.04 & 95.27 & 0.59 & 0.42 \\ 
		    MoCo v2 + CutOut & ResNet-50 & 24 & 256 & 200 & 82.64 & 95.84 & 0.67 & 0.36 \\ 
		    MoCo v2 + CutMix & ResNet-50 &  24 & 256 & 200 & 83.51 & 96.51 & 0.61 & \textbf{0.53} \\ 
			MoCo v2 + MixUp & ResNet-50 & 24 & 256 & 200 & 84.08 & 96.79 & \underline{0.69} & \underline{0.45} \\  \hline
			MoCo v2 + 10\% label & ResNet-50 & 24 & 256 & 200  & 82.26 & 95.80 & \textbf{0.72} & 0.23 \\
			MoCo v2 + 30\% label & ResNet-50 & 24 & 256 & 200  & 82.55 & 95.83 & \textbf{0.72} & 0.23 \\
			MoCo v2 + 50\% label & ResNet-50 & 24 & 256 & 200  & 83.21 & 96.36 & \textbf{0.72} & 0.23 \\
			MoCo v2 + 70\% label & ResNet-50 & 24 & 256 & 200  & 83.75 & 96.62 & \textbf{0.72} & 0.23 \\
			MoCo v2 + 100\% label & ResNet-50 & 24 & 256 & 200 & 84.93 & 97.18 & \textbf{0.72} & 0.23 \\\hline
			MoCo v2 + MixUp + 50\% label & ResNet-50 & 24 & 256 & 200 & \underline{85.59} & \underline{97.43} & \underline{0.69} & \underline{0.45} \\
			MoCo v2 + MixUp + 100\% label & ResNet-50 & 24 & 256 & 200 & \textbf{87.86} & \textbf{98.15} & \underline{0.69} & \underline{0.45} \\
			\bottomrule
			\end{tabular}}
	\label{tab: exp_100}
 \vspace{-1.0em}
\end{table*}

\vspace{-0.5em}
\section{Experiments}
\vspace{-0.5em}
In this part, we conduct extensive experiments by transferring our model to four main downstream tasks, including linear classification, object detection, instance segmentation and semantic segmentation. 
In the meanwhile, we introduce $\mathcal{L}_{inv}$ and $\mathcal{L}_{div}$ to quantify how data augmentation affects the self-/semi-/fully-supervised pre-trained models. 
We give a comprehensive analysis on the effect of data augmentation and supervision during pre-training on various downstream tasks.

\noindent\textbf{Linear Classification.}
Table~\ref{tab: exp_100} reports the top-1 and top-5 accuracy for linear classification on ImageNet-100 benchmark by applying four data augmentations to MoCo v2, where models are trained on frozen features from the pre-trained models.
We can observe that MoCo v2+MixUp achieves better performance than other three data augmentations, including Random Erasing, CutOut, and CutMix.
This is because the augmented views generated from MixUp have larger invariance between themselves and the anchor image.
Meanwhile, with the increase of the number of given labels, we can observe an obvious performance gain in terms of both top-1 and top-5 accuracies, although our augmented views are not changed.
This demonstrates the effectiveness of semi-/fully-supervised learning in learning more meaningful features for classification.
Adding MixUp to the fully-supervised learning boosts the top-1 and top-5 accuracies to 87.86\% and 98.15\%.
In terms of the invariance and diversity between augmented views, adding MixUp to the original MoCo v2 achieves the largest invariance score $\mathcal{L}_{inv}$ with best linear classification performance compared to other data augmentation techniques.
In the meanwhile, all data augmentation techniques indeed increase the diversity score $\mathcal{L}_{div}$ while achieving better results than the baseline, which demonstrates the importance of measuring the quality of the augmented view by the proposed metrics.
Furthermore, adding semi-supervised samples to MoCo v2 do not change the invariance and diversity scores as only augmented views are evaluated during training. 

We compare data augmentation based semi-/fully-supervised models and other self-supervised methods for the linear classification evaluation on ImageNet-1K, as shown in Table~\ref{tab: exp_1k} in Appendix.
Applying MixUp to MoCo v2 increases the top-accuracy from 67.5\% to 68.4\%, which shows the effectiveness of additional data augmentations on the views generated by the baselines.
With the increase of the number of given labels during pre-training, the linear classification accuracy consistently increases.
Particularly, MoCo v2+MixUp+100\% label achieves the best top-1 accuracy in terms of linear classification. 
Please see more experimental details and results in Appendix.

\vspace{-0.5em}
\section{Conclusion}
\vspace{-0.5em}
 
In this work, we perform a comprehensive empirical study to quantify how the self-/semi-/fully- supervised pre-trained models are affected by different data augmentation techniques.
An approach is introduced to measure the quality of the augmented view by explicitly examining the invariance and diversity metrics for self-/semi-/fully- supervised pre-trained models. 
We also conduct extensive experiments on various downstream benchmarks, which demonstrate that invariance and diversity are important metrics for contrastive learning frameworks.
Data augmentations that provide better invariance and diversity result in better performance in downstream tasks.

\newpage

{\small
\bibliography{references}
\bibliographystyle{unsrt}
}

\newpage

\appendix

\section*{Appendix}

\section{Related Work}

\noindent\textbf{Data Augmentation.}
In the vision community, a branch of data augmentation methods~\cite{zhong2020random,devries2017improved,zhang2018mixup,yun2019cutmix} have achieved promising performance in image related tasks, such as image classification and object detection.
Typically, 
Random Erasing~\cite{zhong2020random} selected a rectangle region in an image and erased its pixels with random values to reduce over-fitting and increase the robustness of trained model to occlusion.
CutOut~\cite{devries2017improved} randomly masked square regions of training images and tried to capture less prominent features for classification.
MixUp~\cite{zhang2018mixup} applied a convex combination of pairs of examples and their labels to improve the generalization of neural network architectures.
CutMix~\cite{yun2019cutmix} cut patches and pasted them from training images with mixed ground truth labels to train strong classifiers with localizable features.
Recently, pretext tasks~\cite{zhang2016colorful,yamaguchi2019multiple,deepak2016context,misra2020self,noroozi2016unsupervised,xiao2021what} have been proven to be effective in self-supervised learning for meaningful visual representations. 
Researchers explore various pretext tasks to improve the quality of pre-trained representations, which includes colorization~\cite{zhang2016colorful,yamaguchi2019multiple}, context autoencoders~\cite{deepak2016context}, spatial jigsaw puzzles~\cite{misra2020self,noroozi2016unsupervised} and discriminate orientation~\cite{xiao2021what}.

However, a comprehensive recipe for data augmentations used in self-supervised learning is unexplored.
In this work, we conduct an empirical study to exploit four main data augmentations over self-supervised methods on commonly-used benchmarks in terms of various downstream tasks.
We further introduce invariance and diversity to quantify how data augmentation affects the performance of self-supervised pre-trained models.

\noindent\textbf{Self-Supervised Learning.}
In the self-supervised literature, researchers aim to exploit the internal characteristics of data and leverage pretext tasks to train a model.
Recently, an unsupervised framework that learns effective views with data augmentation was proposed by Tian \textit{et al.}~\cite{tian2020what} to reduce the mutual information between views.
CMC~\cite{tian2020contrastive} introduced a multi-view contrastive learning framework with any number of views to learn view-agnostic representations.
Another pretext task of solving jigsaw puzzles was developed in PIRL~\cite{misra2020self} to improve the semantic quality of learned image representations, achieving better object detection results than supervised pre-training. 

In the past years, contrastive learning has shown its effectiveness in self-supervised learning, where various instance-wise contrastive learning frameworks~\cite{chen2020simple,chen2020big,grill2020bootstrap,he2019moco,chen2020mocov2,khosla2020sup,cao2020parametric,hu2021adco,chen2021simsiam,mo2023mcvt,mo2023contralord} and prototype-level contrastive methods~\cite{caron2020unsupervised,li2021prototypical,wang2021cld,mo2021spcl,mo2022pauc} were proposed.
The general idea of the instance-wise contrastive learning is to close the distance of the embedding of different views from the same instance while pushing embeddings of views from different instances away.
One common way is to use a large batch size to accumulate positive and negative pairs in the same batch. 
For instance, Chen \textit{et al.}~\cite{chen2020simple} proposed a simple framework with a learnable nonlinear projection head and a large batch size to improve the quality of the pre-trained representations.
To make best use of a large amount of unlabelled data, they present a bigger unsupervised pre-training network and introduce distillation with unlabeled data in SimCLR v2~\cite{chen2020big} to improve the performance in downstream tasks. 
The dynamic dictionary was used with a moving-averaged encoder in MoCo series~\cite{chen2020mocov2,he2019moco} to build a dynamic dictionary to to update negative instances in a queue of large size.

Nevertheless, how to leverage labels in the momentum queue based pre-training is unexplored, especially their impacts on various downstream tasks, such as image classification, object detection, and semantic segmentation.
This motivates us to comprehensively explore the effect of self-/semi-/full supervision on pre-trained models that are transferred to the aforementioned tasks. 
In the meanwhile, we quantify the effect of data augmentation on self-/semi-/fully- supervised contrastive learning frameworks.

\section{Pre-training Datasets \& Settings}

Following previous methods~\cite{he2019moco,chen2020mocov2,tian2019contrastive,tian2020contrastive}, we use two popular benchmarks, \textit{ImageNet-100}~\cite{tian2020contrastive} and \textit{ImageNet-1K}.
The ImageNet-100 pre-trained model is evaluated on linear classification, and the ImageNet-1K model is transferred to various downstream tasks, including linear classification, object detection, instance segmentation and semantic segmentation.

For self-supervised pre-training on ImageNet-100 and ImageNet-1K, we closely follow the original MoCo v2 implementation~\cite{chen2020mocov2}.
SGD is used as our optimizer, where we apply a weight decay of 0.0001, a momentum of 0.9, and a batch size of 256.
Our model is trained for 200 epochs with a initial learning rate of 0.03.
The learning rate is then decayed by a factor of 10 at 120 and 160 epochs.
For semi-/fully supervised pre-training, we use the same setting except that some or all labels are provided for maintaining the negative queue with labels.

\section{Transferring Datasets \& Settings}

\noindent \textbf{Linear Classification.}
We evaluate linear classification on \textit{ImageNet-100.} and \textit{ImageNet-1K.} dataset, where a linear classifier is trained on frozen features from pre-trained weights.
We report top-1,top-5 accuracy for ImageNet-100, and top-1 accuracy for ImageNet-1K.

\noindent \textbf{Object Detection.}
For a fair comparison with previous work~\cite{he2019moco,chen2020mocov2}, we fine-tune a Faster R-CNN
detector~\cite{ren2015faster} with C4-backbone end-to-end on the
\textit{PASCAL VOC}~\cite{everingham2010voc} 07+12 trainval set and evaluate on the VOC 07
test set.
For \textit{MS-COCO}~\cite{lin2014coco} benchmark, we use the same hyper-parameters in MoCo~\cite{he2019moco}, and fine-tune a Mask R-CNN~\cite{he2017mask} with C4 backbone on the train2017 set with 2x schedule and evaluate on val2017 set.
The COCO box metrics (AP, AP$_{50}$, AP$_{75}$) are reported on both datasets.

\noindent \textbf{Instance Segmentation.}
In terms of instance segmentation, we evaluate our pre-trained models on three popular benchmarks, including \textit{MS-COCO}~\cite{lin2014coco}, \textit{LVIS v1.0}~\cite{agrim2019lvis}, and \textit{Cityscapes}~\cite{cordts2016cityscapes}.
For MS-COCO, we follow the same setting as the Mask R-CNN~\cite{he2017mask} used in the object detection task, where the COCO mask metrics (AP$^m$, AP$_{50}^m$, AP$_{75}^m$) are reported.
For LVIS, we fine-tune an FCN model~\cite{long2015fcn} on train set for 80k iterations and test on val set.
We use the commonly-used metrics, AP, AP$_c$, AP$_f$, and AP$_r$ for evaluation.
For Cityscapes, an FCN model~\cite{long2015fcn} is fine-tuned end-to-end on train\_fine set for 40k iterations and test on val set, where AP$^m$ and AP$_{50}^m$ are reported for comparison.

\noindent \textbf{Semantic Segmentation.}
We use \textit{Cityscapes}~\cite{cordts2016cityscapes} and \textit{ADE20K}~\cite{zhou2017scene,Zhou2018SemanticUO} to evaluate semantic segmentation.
For both benchmarks, we fine-tune an FCN model~\cite{long2015fcn} on the train set for 40k iterations and test on the val set.
Following previous work~\cite{he2019moco}, we report two metrics (mIoU, mIoU$_{sup}$) for Cityscapes and four metrics (mIoU, fwIoU, mACC, pACC) for ADE20K to have a comprehensive comparison.

\section{Additional Experiments}

\begin{table*}[t]
    \centering 
    \caption{Comparisons of linear classification evaluation on ImageNet-1K, where all results are trained under the same architecture. 
    Parameters are of the feature extractor~\cite{he2020momentum}. 
    Views denote the number of images fed into the encoder in one iteration under batch size 1.}
    \renewcommand\tabcolsep{3.0pt}
	\centering
 \vspace{0.5em}
	\scalebox{0.85}{
    \begin{tabular}{lccccccc}
    \toprule
    Method & Arch. & Param.(M) & Batch & Epochs &  Views & Top-1 (\%)  \\ \midrule
        NPID~\cite{wu2018unsupervised} & ResNet-50 & 24 & 256 & 200   & 2x224 & 58.5   \\ 
        LocalAgg~\cite{zhuang2019local} & ResNet-50 & 24 & 128 & 200   & 2x224 & 58.8  \\ 
        MoCo~\cite{he2019moco} & ResNet-50 & 24 & 256 & 200  & 2x224 & 60.6 \\ 
        SimCLR~\cite{chen2020simple} & ResNet-50 & 24  & 256 & 200 & 2x224 & 61.9 \\
        CPC v2~\cite{oord2018representation} & ResNet-50 & 24 & 512  & 200  & 2x224 & 63.8  \\
        CMC~\cite{tian2020contrastive} & ResNet-50 & 47 & 128 & 240 & 2x224 & 66.2   \\
        MoCo v2~\cite{chen2020mocov2} & ResNet-50 & 24 & 256 & 200   & 2x224 & 67.5 \\
        PCL v2~\cite{li2021prototypical} & ResNet-50 & 24  & 512 & 200 & 2x224 &  67.6 \\ 
        PIC~\cite{cao2020parametric} & ResNet-50 & 24  & 512 & 200  & 2x224 & 67.6 \\ 
        MoCHi~\cite{kalantidis2020hard} & ResNet-50 & 24 & 512 & 200  & 2x224 & 68.0 \\ 
        AdCo~\cite{hu2021adco} & ResNet-50 & 24 & 256 & 200 & 2x224 & 68.6 \\ 
        SwAV~\cite{caron2020unsupervised} & ResNet-50 & 24 & 4096 & 200 & 2x224 & 69.1 \\
        LoCo~\cite{xiong2020loco} & ResNet-50 & 24 & 4096 & 800 & 2x224 & 69.5 \\
        BYOL~\cite{grill2020bootstrap} & ResNet-50 & 24 & 4096 & 200 & 4x224 &  70.6 \\
        SimSiam~\cite{chen2021simsiam} & ResNet-50 & 24 & 256 & 200 & 4x224 &  70.0 \\ \midrule
        MoCo v2 + MixUp & ResNet-50 & 24 & 256 & 200   & 2x224 & 68.4 \\ 
        MoCo v2 + MixUp + 50\% label & ResNet-50 & 24 & 256 & 200   & 2x224 & 69.3 \\ 
			MoCo v2 + MixUp + 100\% label & ResNet-50 & 24 & 256 & 200   & 2x224 & \textbf{71.2} \\  
	    \bottomrule
    \end{tabular}}
    \label{tab: exp_1k}  
\vspace{-1.0em}
\end{table*}

\noindent \textbf{Object Detection.}
We transfer various self-supervised pre-trained models to PASCAL VOC for object detection, and report the comparison results of AP, AP$_{50}$, and AP$_{75}$ in Table~\ref{tab: obj_voc}.
As can be seen, adding MixUp to the pre-training with the highest invariance achieves the best results compared to other data augmentations. 
This further shows the importance of learning the invariance during pre-training for object detection on PASCAL VOC.
We further evaluate our models pre-trained by various data augmentations on MS-COCO for a comprehensive comparison.
The experimental results are reported in Table~\ref{tab: obj_coco}.
MoCo v2 + MixUp consistently achieves the best performance in terms of all metrics (AP$^b$, AP$_{50}^b$, AP$_{75}^b$), which further demonstrates the effectiveness of MixUp in learning a larger invariance between the augmented views and the anchor image.

\newcommand{\voc}{
\begin{tabular}{llll}
			\toprule
            Method & AP & $AP_{50}$ & $AP_{75}$ \\
			\midrule
			Random Initialization  & 32.80 & 59.00 & 31.60 \\
			Supervised & 54.20 & 81.60 & 59.80 \\
			SimCLR~\cite{chen2020simple} & 51.50 & 79.40 & 55.60 \\
			BOYL~\cite{grill2020bootstrap} & 51.90 & 81.00 & 56.50 \\
        	SwAV~\cite{caron2020unsupervised} & 55.40 & 81.50 & 61.40 \\
			MoCo~\cite{he2019moco} & 55.90 & 81.50 & 62.60 \\
			MoCov2~\cite{chen2020mocov2} & 57.00 & 82.40 & 63.60 \\
    	   SimSiam~\cite{chen2021simsiam} & 57.00 & 82.40 & 63.70 \\ \midrule
    	   MoCov2 + Random Erasing & 56.39 & 81.79 & 62.92 \\
    	    MoCov2 + CutOut        & 57.49 & 82.83 & 63.06 \\
    	    MoCov2 + CutMix        & \underline{57.22} & \underline{82.91} & \underline{63.95} \\
    	     MoCov2 + MixUp        & \textbf{57.61} & \textbf{82.96} & \textbf{64.30} \\			
			\bottomrule
			\end{tabular}
}

\newcommand{\coco}{
\begin{tabular}{lllllll}
			\toprule
            Method & AP$^b$ & AP$_{50}^b$ & AP$_{75}^b$ & AP$^m$ & AP$_{50}^m$ & AP$_{75}^m$ \\
			\midrule
			Random Initialization  & 32.80 & 50.90 & 35.30 & 29.90 & 47.90 & 32.00  \\
			Supervised~ & 39.70 & 59.50 & 43.30 & 35.90 & 56.60 & 38.60 \\
			SwAV~\cite{caron2020unsupervised} & 37.60 & 57.60 & 40.30 & 33.10 & 54.20 & 35.10 \\
			SimSiam~\cite{chen2021simsiam} & 39.20 & 59.30 & 42.10 & 34.40 & 56.00 & 36.70 \\
			MoCo~\cite{he2019moco} & 40.70 & 60.50 & 44.10 & 35.40 & 57.30 & 37.60 \\
			MoCHi~\cite{kalantidis2020hard} & 39.40 & 59.00 & 42.70 & 34.50 & 55.70 & 36.70 \\
			MoCov2~\cite{chen2020mocov2} & 39.80 & 59.80 & \underline{43.60} & \textbf{36.10} & 56.90 & \underline{38.70} \\
			PCL~\cite{li2021prototypical} & \underline{41.00} & \underline{60.80} & 44.20 & 35.60 & 57.40 & 37.80 \\ \midrule
			MoCov2 + Random Erasing & 40.14 & 60.25 & 43.82 & 35.35 & 57.13 & 37.75 \\
    	    MoCov2 + CutOut         & 40.84 & 60.73 & 44.25 & 35.72 & \underline{57.41} & \textbf{38.76} \\
    	    MoCov2 + CutMix         & 40.75 & 60.67 & 44.12 & 35.53 & 57.23 & 38.24 \\
    	     MoCov2 + MixUp         & \textbf{41.07} & \textbf{60.96} & \textbf{44.50} & \underline{36.05} & \textbf{57.69} & 38.37 \\
			\bottomrule
			\end{tabular}
}

\begin{table}[t!]
    \centering
    \caption{Comparison results of object detection and instance segmentation on PASCAL VOC \& COCO. 
    Bold and underline denote the first and second place.
    \label{tab: exp_objdet}}
    \vspace{0.3em}
    \begin{subfigure}{0.4\textwidth}
        \resizebox{\linewidth}{!}{\voc}
        \caption{PASCAL VOC.}
       \label{tab: obj_voc}
    \end{subfigure}\hfill
    \begin{subfigure}{0.58\textwidth}
        \resizebox{\linewidth}{!}{\coco}
        \caption{COCO.}
        \label{tab: obj_coco}
    \end{subfigure}\hfill
    \vspace{-1.5em}
\end{table}

\noindent \textbf{Instance Segmentation.}
The comparison results of instance segmentation on MS-COCO are reported in Table~\ref{tab: obj_coco}.
We can observe that MoCov2 + CutOut achieves the best AP$_{75}^m$ compared to other data augmentations. 
This is because MoCov2 + CutOut has the lowest diversity $\mathcal{L}_{div}$ between augmented views, demonstrating the importance of reducing the diversity of augmented views to improve the performance of instance segmentation.
In Table~\ref{tab: seg_lvis}, we report the comparison results of instance segmentation by fine-tuning our pre-trained models on LVIS v1.0 benchmark.
MoCo v2 + MixUp outperforms MoCo v2 + CutOut by a small margin since they achieves comparable diversity score $\mathcal{L}_{div}$ between augmented views, as we reported in Table~\ref{tab: exp_100}.   
Moreover, MoCo v2 + Random Erasing achieves the worst performance in terms of all metrics.
This shows the importance of keeping invariant features during pre-training while increasing the diversity of augmented views.
We compare the results of instance segmentation on Cityscapes in Table~\ref{tab: seg_cityscapes}.
We can observe a similar trend as LVIS v1.0 dataset, where MoCo v2 + MixUp performs the best while MoCo v2 + Random Erasing performs the worst, which further demonstrates the importance of learning the invariances from augmented views during pre-training and increasing the diversity of augmented views at the same time.

\newcommand{\lvis}{
\begin{tabular}{lcccc}
			\toprule
Method & AP & AP$_c$ & AP$_f$ & AP$_r$  \\
			\midrule
			MoCov2~\cite{chen2020mocov2} & 17.08 & 8.16 & 15.35 & 22.94 \\
            + Random Erasing & 16.92 & 8.03 & 15.12 & 22.85 \\ 
    	    + CutOut         & \underline{17.19} & \underline{8.18} & \underline{15.36} & \underline{23.06} \\ 
    	    + CutMix         & 17.11 & 8.16 & 15.35 & 22.96 \\ 
    	    + MixUp         & \textbf{17.33} & \textbf{8.22} & \textbf{15.42} & \textbf{23.09} \\ 
			\bottomrule
			\end{tabular}
}

\newcommand{\cityscapes}{
\begin{tabular}{lcccc}
			\toprule
Method & AP$^{m}$ & AP$^{m}_{50}$ & mIoU & mIoU$_{sup}$ \\
			\midrule
			MoCov2~\cite{chen2020mocov2} & 22.57 & 48.19 & 55.48 & 79.72 \\
            + Random Erasing & 22.51 & 48.15 & 55.35 & 79.63 \\
    	    + CutOut          & \underline{22.76} & \underline{48.25} & \underline{55.80} & \underline{79.90} \\
    	    + CutMix          & 22.55 & 48.19 & 55.45 & 79.67 \\
    	    + MixUp          & \textbf{22.83} &\textbf{48.28} & \textbf{55.92} & \textbf{79.96} \\
			\bottomrule
			\end{tabular}
}

\newcommand{\ade}{
\begin{tabular}{lcccc}
			\toprule
Method & mIoU & fwIoU & mACC & pACC \\
			\midrule
			MoCov2~\cite{chen2020mocov2} & 20.62 & 54.68 & 27.15 & 69.59 \\
            + Random Erasing & 20.51 & 54.61 & 27.07 & 69.52 \\
    	    + CutOut          & \underline{20.76} & \underline{54.72} & \underline{27.19} & \underline{69.61} \\
    	    + CutMix          & 20.67 & 54.69 & 27.09 & 69.55 \\
    	    + MixUp          & \textbf{20.93} & \textbf{54.80} & \textbf{27.26} & \textbf{69.65} \\
			\bottomrule
			\end{tabular}
}

\begin{table}[t!]
    \centering
    \caption{Comparison results of instance and semantic segmentation. 
    Bold and underline denote the first and second place.
    \label{tab: exp_seg}}
    \vspace{0.3em}
    \begin{subfigure}{0.31\textwidth}
        \resizebox{\linewidth}{!}{\lvis}
        \caption{LVIS.}
        \label{tab: seg_lvis}
    \end{subfigure}\hfill
    \begin{subfigure}{0.33\textwidth}
        \resizebox{\linewidth}{!}{\cityscapes}
        \caption{Cityscapes.}
        \label{tab: seg_cityscapes}
    \end{subfigure}\hfill
    \begin{subfigure}{0.33\textwidth}
        \resizebox{\linewidth}{!}{\ade}
        \caption{ADE20K.}
        \label{tab: seg_ade}
    \end{subfigure}\hfill
    \vspace{-2.0em}
\end{table}

\noindent \textbf{Semantic Segmentation.}
Table~\ref{tab: seg_cityscapes} shows the comparison results of semantic segmentation fine-tuned on Cityscapes dataset.
MoCov2 + MixUp and MoCov2 + CutOut achieve comparable performance in terms of both metrics.
This shows the effectiveness of learning the invariance and diversity together from augmented views during pre-training.
With the smallest invariance score $\mathcal{L}_{inv}$, MoCov2 + Random Erasing performs worse than other data augmentations.
In Table~\ref{tab: seg_ade}, we report the comparison results of semantic segmentation fine-tuned on ADE20K dataset.
We can make similar observations as the Cityscapes dataset.
Compared to other data augmentations, MoCo v2 + Random Erasing achieves the worst results while MoCov2 + MixUp achieves the best performance.
This further demonstrates the effectiveness of MixUp in keeping the invariance and increasing the diversity at the pre-training stage.

\section{Additional Analysis}

In this part, we explore the effect of the number of augmented views $V$ and batch size $N$ on the invariance and diversity.
All experiments for ablation studies are conducted with MoCo v2 + MixUp on ImageNet-100 dataset.

\noindent\textbf{Number of augmented views.}
In order to explore how the number of augmented $V$ views affects the invariance and diversity, we set the value of $V$ to 2, 3, and 4.
The experimental results are reported in Table~\ref{tab: ab_aug}.
As can be seen, when $V$ is set to 2, we achieve the best top-1 and top-5 accuracies with the largest invariance score $\mathcal{L}_{inv}$ and the smallest diversity score $\mathcal{L}_{div}$.
With the increase in the number of augmented views, the performance of our model decreases a lot, which demonstrates the importance of selecting the right augmented views for contrastive learning.

\noindent\textbf{Batch size.}
In order to demonstrate the effect of batch size on the final performance of invariance and diversity.
Specifically, we set the number of batch size $N$ to 32, 64, 128, 256, 512, 1024, and report the comparison results in Table~\ref{tab: ab_bs}.
When the batch size is set to 256, our model achieves the best performance in terms of the top-1 and top-5 accuracy.
In the meanwhile, with the decrease in the batch size, both the invariance and diversity score increases, resulting in performance degradation.

\newcommand{\aug}{
\begin{tabular}{ccccc}
		\toprule
	    \# of views ($V$) & Top-1 (\%) & Top-5 (\%) & $\mathcal{L}_{inv}$ &  $\mathcal{L}_{div}$ \\
	    	
		\midrule
		2 & \textbf{84.08} & \textbf{96.79} & \textbf{0.69} & 0.45 \\
		3 & 82.37 & 95.81 & 0.58 & 0.53 \\
		4 & 81.55 & 95.68 & 0.51 & \textbf{0.59} \\
		\bottomrule
\end{tabular}
}

\newcommand{\bs}{
\begin{tabular}{ccccc}
		\toprule
	    batch size ($N$) & Top-1 (\%) & Top-5 (\%) & $\mathcal{L}_{inv}$ &  $\mathcal{L}_{div}$ \\
	    	
		\midrule
		32 & 82.13 &  95.65 & \textbf{0.75} & 0.52 \\ 
		64 & 82.78 &  95.91 & 0.73 & 0.49 \\ 
		128 & 83.27 &  96.38 & 0.72 & 0.47 \\ 
		256 & \textbf{84.08} &  \textbf{96.79} & 0.69 & \textbf{0.45} \\ 
		512 & 83.49 &  96.52 & 0.61 & 0.57 \\ 
		1024 & 82.92 &  96.23 & 0.58 & 0.63 \\ 
		\bottomrule
\end{tabular}
}

\begin{table}[t!]
    \centering
    \caption{Ablation Studies on augmented views and batch size, where top-1, top-5 accuracy, $\mathcal{L}_{inv}$, and $\mathcal{L}_{div}$ are reported on ImageNet-100.
    \label{tab: exp_ab}}
    \vspace{0.3em}
    \begin{subfigure}{0.5\textwidth}
        \resizebox{\linewidth}{!}{\aug}
        \caption{Augmented Views.}
       \label{tab: ab_aug}
    \end{subfigure}\hfill
    \begin{subfigure}{0.45\textwidth}
        \resizebox{\linewidth}{!}{\bs}
        \caption{Batch Size.}
       \label{tab: ab_bs}
    \end{subfigure}\hfill
    \vspace{-2.0em}
\end{table}

\section{Limitation} 
The crucial limitation of this work is the scale of the datasets and backbones. Due to limited computational resources, the majority of the experiments are carried out on the ImageNet-100 dataset using the ResNet-50. Therefore we are unsure about the availability of the conclusions on much larger datasets and backbones. For instance, we do not perform experiments on costful transformer-based frameworks, such as DINO~\cite{caron2021emerging}. 
Nevertheless, we consider the results should generalize to other situations. 
On the other hand, we cannot enumerate all types of data augmentations that mask out information about the image. In recent studies, the patch-wise CutOut is shown effective in self-supervised algorithms such as masked image modeling. While in this work, we focus on the contrastive learning algorithm, the analysis of other data augmentations will be conducted in future works.

\section{Broader Impact.} 

The empirical results of our study benefit self-/semi-/fully- supervised pre-trained frameworks in the literature. 
Moreover, the analysis of the invariance and diversity terms helps in designing the appropriate data augmentation for the downstream tasks.

\end{document}